\documentclass[letterpaper]{article}

\usepackage{amssymb}
\usepackage{graphicx}

\newcommand{\Scal}{\mathcal{S}}
\newcommand{\Vcal}{\mathcal{V}}
\newcommand{\Ecal}{\mathcal{E}}
\newcommand{\Kcal}{\mathcal{K}}

\begin{document}

\title{A Keygraph Classification Framework\\
for Real-Time Object Detection}

\author{Marcelo Hashimoto \and Roberto M. Cesar Jr.}

\maketitle

\begin{abstract}
  In this paper, we propose a new approach for keypoint-based object
  detection. Traditional keypoint-based methods consist in classifying
  individual points and using pose estimation to discard
  misclassifications. Since a single point carries no relational
  features, such methods inherently restrict the usage of structural
  information to the pose estimation phase. Therefore, the classifier
  considers purely appearance-based feature vectors, thus requiring
  computationally expensive feature extraction or complex
  probabilistic modelling to achieve satisfactory robustness. In
  contrast, our approach consists in classifying graphs of keypoints,
  which incorporates structural information during the classification
  phase and allows the extraction of simpler feature vectors that are
  naturally robust. In the present work, 3-vertices graphs have been
  considered, though the methodology is general and larger order
  graphs may be adopted. Successful experimental results obtained for
  real-time object detection in video sequences are reported.
\end{abstract}

\section{Introduction}

\emph{Object detection} is one of the most classic problems in
computer vision and can be informally defined as follows: given an
image representing an object and another, possibly a video frame,
representing a scene, decide if the object belongs to the scene and
determine its \emph{pose} if it does. Such pose consists not only of
the object location, but also of its scale and rotation. The object
might not even be necessarily rigid, in which case more complex
deformations are possible. We will refer to the object image as our
\emph{model} and, for the sake of simplicity, refer to the scene image
simply as our \emph{frame}.

Recent successful approaches to this problem are based on
\emph{keypoints}
\cite{lowe:ijcv-20-91-2004,btv:p9eccv-404-2006,lf:itpami-28-1465-2006,ofl:picsccvpr-1-2007}. In
such approaches, instead of the model itself, the algorithm tries to
locate a subset of points from the object. The chosen points are those
that satisfy desirable properties, such as ease of detection and
robustness to variations of scale, rotation and brightness. This
approach reduces the problem to \emph{supervised classification} where
each model keypoint represents a class and feature vectors of the
frame keypoints represent input data to the classifier.

A well-known example is the SIFT method proposed by Lowe
\cite{lowe:ijcv-20-91-2004}. The most important aspect of this method
relies on the very rich feature vectors calculated for each keypoint:
they are robust and distinctive enough to allow remarkably good
results in practice even with few vectors per class and a simple
nearest-neighbor approach. More recent feature extraction strategies,
such as the SURF method proposed by Bay, Tuytelaars and van Gool
\cite{btv:p9eccv-404-2006}, are reported to perform even better.

The main drawback of using rich feature vectors is that they are
usually complex or computationally expensive to calculate, which can
be a shortcoming for real-time detection in videos, for
example. Lepetit and Fua \cite{lf:itpami-28-1465-2006} worked around
this limitation by shifting much of the computational burden to the
training phase. Their method uses simple and cheap feature vectors,
but extracts them from several different images artificially generated
by applying changes of scale, rotation and brightness to the
model. Therefore, robustness is achieved not by the richness of each
vector, but by the richness of the training set as a whole.

Regardless the choice among most of such methods, keypoint-based
ap\-proach\-es traditionally follow the same general framework,
described in Figure~\ref{fig:original}.

\begin{figure}
\centering
\footnotesize
\begin{tabular}{|p{300px}c|}
\hline
\textbf{Training}&~~\\
\hline
\begin{enumerate}
\item Detect keypoints in the model.

\item Extract feature vectors from each keypoint.

\item Use the feature vectors to train a classifier whose classes are
  the keypoints. The accuracy must be reasonably high, but not
  necessarily near-perfect.
\end{enumerate}&\\
\hline
\textbf{Classification}&\\
\hline
\begin{enumerate}
\item Detect keypoints in the frame.
\item Extract feature vectors from each keypoint.

\item Apply the classifier to the feature vectors in order to decide
  if each frame keypoint is sufficiently similar to a model keypoint.
  As near-perfect accuracy is not required, several misclassifications
  might be done in this step.

\item Use an estimation algorithm to determine a pose spatially
  coherent with a large enough number of classifications made during
  the previous step. Classifications disagreeing with such pose are
  discarded as outliers.
\end{enumerate}&\\
\hline
\end{tabular}
\normalsize
\caption{\label{fig:original} Traditional framework for keypoint-based
  object detection.}
\end{figure}

A shortcoming of this framework is that structural information, such
as geometric and topological relations between the points, only play a
role in the pose estimation step: they are completely absent of all
classification steps. Therefore, the entire burden of describing a
keypoint lies on individual appearance information, such as the color
of pixels close to it. Recently, \"Ozuysal, Fua and
Lepetit~\cite{ofl:picsccvpr-1-2007} proposed a less individual
approach by defining a probabilistic modelling scheme where small
groups of keypoints are considered. However, since a purely
appearance-based feature vector set is used under this model, there is
still an underuse of structure in their approach.

In this paper, we propose an alternative framework that, instead of
classifying single keypoints, classifies sets of keypoints using both
appearance and structural information. Since graphs are mathematical
objects that naturally model relations, they are adopted to represent
such sets. Therefore, the proposed approach is based on supervised
classification of graphs of keypoints, henceforth referred as
\emph{keygraphs}. A general description of our framework is given
by Figure \ref{fig:framework}.

\begin{figure}
\centering
\footnotesize
\begin{tabular}{|p{300px}c|}
\hline
\textbf{Training}&~~\\
\hline
\begin{enumerate}
\item Detect keypoints in the model.
\item \emph{Build a set of keygraphs whose vertices are the detected
    keypoints.}

\item Extract feature vectors from each \emph{keygraph}.

\item Use the feature vectors to train a classifier whose classes are
  the \emph{keygraphs}. The accuracy must be reasonably high, but not
  necessarily near-perfect.
\end{enumerate}&\\
\hline
\textbf{Classification}&\\
\hline
\begin{enumerate}
\item Detect keypoints in the frame.
\item \emph{Build a set of keygraphs whose vertices are the detected
    keypoints.}

\item Extract feature vectors from each \emph{keygraph}.

\item Apply the classifier to the feature vectors in order to decide
  if each frame \emph{keygraph} is sufficiently similar to a model
  \emph{keygraph}.  As near-perfect accuracy is not required, several
  misclassifications might be done in this step.

\item Use an estimation algorithm to determine a pose spatially
  coherent with a large enough number of classifications made during
  the previous step. Classifications disagreeing with such pose are
  discarded as outliers.
\end{enumerate}&\\
\hline
\end{tabular}
\normalsize
\caption{\label{fig:framework} Proposed framework with the main
  differences emphasized.}
\end{figure}

The idea of using graphs built from keypoints to detect objects is not
new: Tang and Tao \cite{tt:2jiiwvspets-1-2005} had success with
dynamic graphs defined over SIFT points. Their work, however, shifts
away from the classification approach and tries to solve the problem
with \emph{graph matching}. Our approach, in contrast, still reduces
the problem to supervised classification, which is more efficient. In
fact, it can be seen as a generalization of the traditional methods,
since a keypoint is a single-vertex graph.

This paper is organized as follows. Section 2 introduces the proposed
framework, focusing on the advantages of using graphs instead of
points. Section 3 describes a concrete implementation of the
framework, where 3-vertices keygraphs are used, and some successful
experimental results that this implementation had for real-time object
detection.  Finally, in Section 4 we present our conclusions.

\section{Keygraph Classification Framework}

A \emph{graph} is a pair~$(\Vcal, \Ecal)$, where~$\Vcal$ is an
arbitrary set, $\Ecal \subseteq {\Vcal \choose 2}$ and~$\Vcal \choose
2$ denotes the family of all subsets of~$\Vcal$ with
cardinality~$2$. We say that~$\Vcal$ is the set of \emph{vertices}
and~$\Ecal$ is the set of \emph{edges}. We also say that the graph is
\emph{complete} if~$\Vcal = {\Vcal \choose 2}$ and that~$(\Vcal',
\Ecal')$ is a \emph{subgraph} of~$(\Vcal, \Ecal)$ if~$\Vcal' \subseteq
\Vcal$ and~$\Ecal' \subseteq \Ecal \cap {\Vcal' \choose 2}$. Given a
set~$\Scal$, we denote by~$G(\Scal)$ the complete graph whose set of
vertices is~$\Scal$.

Those definitions allow us to easily summarize the difference between
the traditional and the proposed frameworks. Both have the same
outline: define certain universe sets from the model and the frame,
detect key elements from those sets, extract feature vectors from
those elements, train a classifier with the model vectors, apply the
classifier to the frame vectors and analyze the result with a pose
estimation algorithm. The main difference lies on the first step:
defining the universe set of an image. In the traditional framework,
since the set of keypoints~$\Kcal$ represents the key elements, this
universe is the set of all image points. In the proposed framework,
while the detection of~$\Kcal$ remains, the universe is the set of all
subgraphs of~$G(\Kcal)$. In the following subsections, we describe the
fundamental advantages of such difference in three steps: the key
element detection, the feature vector extraction and the pose
estimation.

\subsection{Keygraph Detection}

\begin{figure}
\centering
\includegraphics{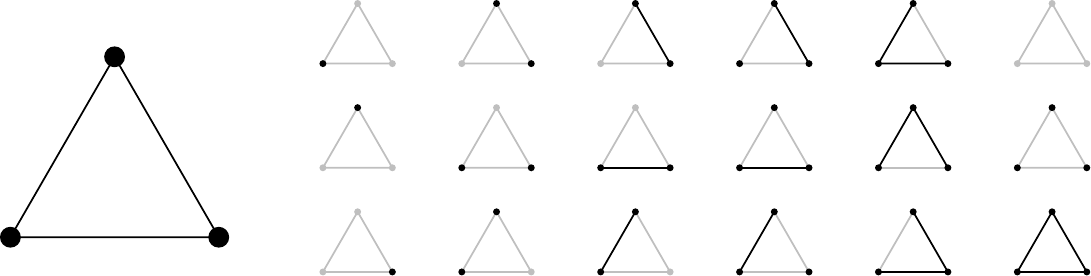}
\caption{\label{fig:subgraphs} Even a very simple graph with three
  vertices has a large number of subgraphs.}
\end{figure}

As it can be seen on Figure~\ref{fig:subgraphs}, one of the most
evident differences between detecting a keypoint and detecting a
keygraph is the size of the universe set: the number of subgraphs
of~$G(\Kcal)$ is exponential on the size of~$\Kcal$.

\begin{figure}
\centering
\includegraphics{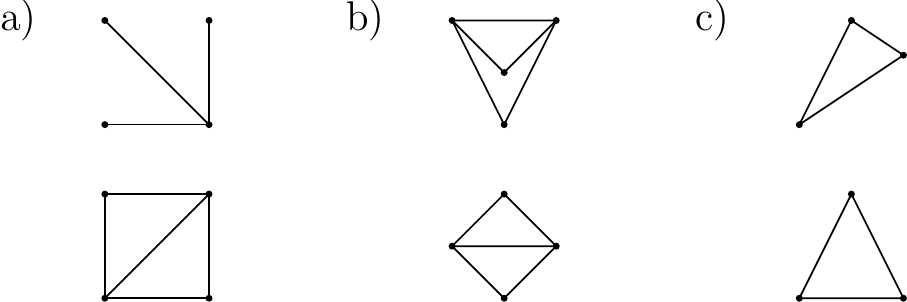}
\caption{\label{fig:structure} Gradual restriction by structural
  properties. Column (a) shows two graphs with different combinatorial
  structure. Column (b) shows two graphs combinatorially equivalent
  but topologically different. Finally, column (c) shows two graphs
  with the same combinatorial and topological structure, but different
  geometric structure.}
\end{figure}

This implies that a keygraph detector must be much more restrictive
than a keypoint detector if we are interested in real-time
performance. Such necessary restrictiveness, however, is not hard to
obtain because graphs have structural properties to be explored that
individual keypoints do not. Those properties can be classified in
three types: \emph{combinatorial}, \emph{topological} and
\emph{geometric}. Figure \ref{fig:structure} shows how those three
types of structural properties can be used to gradually restrict the
number of considered graphs.

\subsection{Partitioning the Feature Vectors}

A natural approach for extracting feature vectors from keygraphs is by
translating all the keygraph properties, regardless if they are
structural or appearance-based, into scalar values. However, a more
refined approach that allows to take more advantage of the power of
structural information has been adopted.

\begin{figure}
\centering
\begin{tabular}{cc}
\includegraphics[scale=0.5]{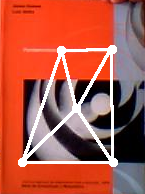}&
\includegraphics{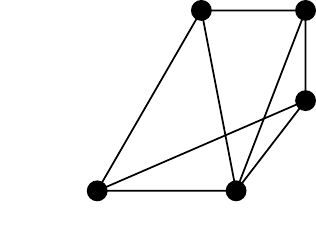}\\
(a)&(b)
\end{tabular}
\caption{\label{fig:impossible} Model keygraph (a) and a frame
  keygraph (b) we want to classify. From the topological structure
  alone we can verify that the latter cannot be matched with the
  former: the right graph does not have a vertex inside the convex
  hull of the others. Furthermore, translating this simple boolean
  property into a scalar value does not make much sense.}
\end{figure}

This approach consists in keeping the feature vectors themselves
appearance-based, but partitioning the set of vectors according to
structural properties. There are two motivations for such approach:
the first one is the fact that a structural property, alone, may
present a strong distinctive power. The second one is the fact that
certain structural properties may assume boolean values for which a
translation to a scalar does not make much sense. Figure
\ref{fig:impossible} gives a simple example that illustrates the two
motivations.

By training several classifiers, one for each subset given by the
partition, instead of just one, we not only satisfy the two
motivations above, but we also improve the classification from both an
accuracy and an efficiency point of view.

\subsection{Naturally Robust Features}

\begin{figure}
\centering
\includegraphics[scale=0.5]{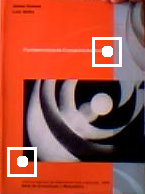}
\hspace*{50pt} \includegraphics[scale=0.5]{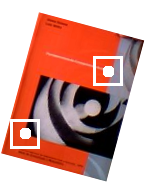}
\caption{\label{fig:patches} Lack of robustness from patch methods.
  The keypoints are the same in both images, but the small change in
  scale and rotation gives completely different patches.}
\end{figure}

For extracting a feature vector from a keygraph, there exists a
natural approach by merging multiple keypoint feature vectors
extracted from its vertices. However, a more refined approach may be
derived. In traditional methods, a keypoint feature vector is
extracted from color values of the points that belong to a certain
\emph{patch} around it. This approach is inherently flawed because, as
Figure~\ref{fig:patches} shows, such patches are not naturally robust
to scale and rotation.

Traditional methods work around this flaw by improving the
extraction itself. Lowe \cite{lowe:ijcv-20-91-2004} uses a gradient
histogram approach, while Lepetit and Fua
\cite{lf:itpami-28-1465-2006} rely on the training with multiple
sintethic views.

\begin{figure}
\centering
\includegraphics[scale=0.5]{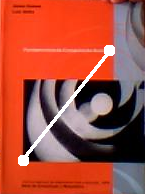}
\hspace*{50pt} \includegraphics[scale=0.5]{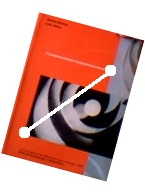}
\caption{\label{fig:relative} An example of relative feature
  extraction. Both keygraphs consist of two keypoints and the edge
  between them. Suppose there is no variation of brightness between
  the two images and consider for each keygraph the mean gray level
  relative to all image pixels crossed by its edge. Regardless of
  scale and rotation, there should be no large variations between the
  two means. Therefore, they represent a naturally robust feature.}
\end{figure}

With keygraphs, in contrast, the flaw does not exist in the first
place, because they are built on \emph{sets} of keypoints. Therefore,
they allow the extraction of \emph{relative} features that are
naturally robust to scale and rotation without the need of
sophisticated extraction strategies. Figure \ref{fig:relative} shows a
very simple example.

\subsection{Pose Estimation by Voting}\label{subsec:voting}

A particular advantage of the SIFT feature extraction scheme relies on
its capability of assigning, to each feature vector, a scale and
rotation relative to the scale and rotation of the model itself. This
greatly reduces the complexity of pose estimation because each
keypoint classification naturally induces a pose that the object must
have in the scene if such classification is correct. Therefore, one
can obtain a robust pose estimation and discard classifier errors by
simply following a Hough \cite{hough:usp-1962} transform procedure: a
quantization of all possible poses is made and each evaluation from
the classifier registers a vote for the corresponding quantized
pose. The most voted pose wins.

\begin{figure}
\centering
\includegraphics[scale=0.5]{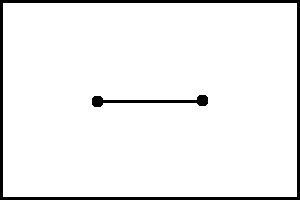}

\vspace*{1pt}

\includegraphics[scale=0.5]{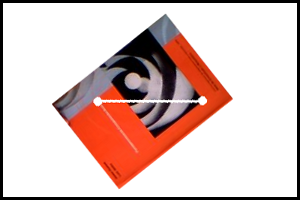}
\includegraphics[scale=0.5]{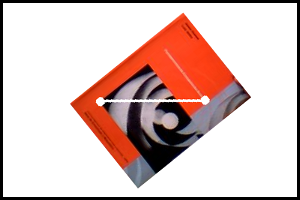}
\caption{\label{fig:poor} Example of pose estimation ambiguity. The
  white rectangle indicates the pose of a certain 2-vertex graph in a
  frame. If a classifier evaluates this graph as being the model
  keygraph indicated in Figure \ref{fig:relative}, there would be two
  possible coherent rotations.}
\end{figure}

The same procedure can be used with keygraphs, because relative
properties of a set of keypoints can be used to infer scale and
rotation. It should be emphasized, however, that the viability of such
strategy depends on how rich the structure of the considered keygraphs
is. Figure \ref{fig:poor} has a simple example of how a poorly chosen
structure can cause ambiguity during the pose estimation.

\section{Implementation and Results}

In this section we will present the details of an implementation of
the proposed framework that was made in C++ with the OpenCV
\cite{opencv} library.  To illustrate our current results with this
implementation, we describe an experiment on which we attempted to
detect a book in real-time with a webcam, while varying its position,
scale and rotation. We ran the tests in an Intel\textregistered{}
Core\texttrademark{} 2 Duo T7250 with 2.00GHz and 2 GB of RAM. A
2-megapixel laptop webcam was used for the detection itself and to
take the single book picture used during the training.

\subsection{Good Features to Track}

\begin{figure}
\centering
\includegraphics[scale=0.5]{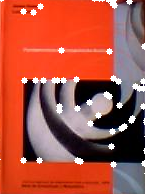}
\caption{\label{fig:keypoints} Result of the \emph{good features to
    track} algorithm.}
\end{figure}

For keypoint detection we used the well-known \emph{good features to
  track} detector proposed by
Shi~and~Tomasi~\cite{st:picsccvpr-593-1994}, that applies a threshold
over a certain quality measure. By adjusting this threshold, we are
able to control how rigid is the detection. A good balance between
accuracy and efficiency was found in a threshold that gave us 79
keypoints in the model, as it can be seen on
Figure~\ref{fig:keypoints}.

\subsection{Thick Scalene Triangles}

For keygraph detection we selected 3-vertices complete graphs whose
induced triangle is sufficiently thick and scalene. More formally,
that means each one of the internal angles in the triangle should be
larger than a certain threshold and the difference between any two
internal angles is larger than another threshold. The rationale behind
this choice is increasing structure richness: the vertices of a
excessively thin triangle are too close of being collinear and high
similarity between internal angles could lead to the pose estimation
ambiguity problem mentioned in the previous section.

In our experiment, we established that no internal angle should have
less than 5 degrees and no pair of angles should have less than 5
degrees of difference. To avoid numerical problems, we also added that
no pair of vertices should have less than 10 pixels of distance. Those
three thresholds limited drastically the number of keygraphs: out of
$79 \cdot 78 \cdot 77 = 474.474$ possible 3-vertices subgraphs, the
detector considered $51.002$ keygraphs.

\begin{figure}
  \centering
  \includegraphics{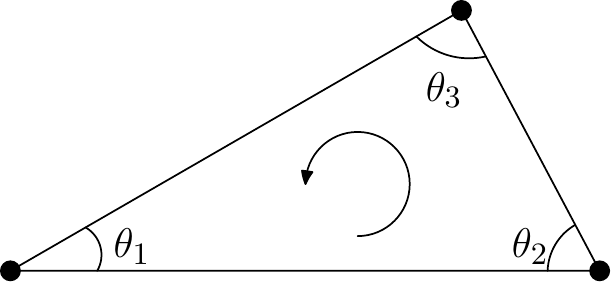}
  \caption{\label{fig:partition} Scalene triangle with $\theta_1 <
    \theta_2 < \theta_3$. In this case, if we pass through the three
    vertices in increasing order of internal angle, we have a
    counter-clockwise movement.}
\end{figure}

The partitioning of the feature vector set is made according to three
structural properties. Two of them are the values of the two largest
angles. Notice that, since the internal angles of a triangle always
sum up to 180 degrees, considering all angles would be redundant. The
third property refers to a clockwise or counter-clockwise direction
defined by the three vertices in increasing order of internal
angle. Figure \ref{fig:partition} has a simple example.

In our experiment we established a partition in~$2 \cdot 36 \cdot 36
= 2592$ subsets: the angles are quantized by dividing the
interval~$(0, 180)$ in~$36$ bins. The largest subset in the partition
has~$504$ keygraphs, a drastic reduction from the~$51.002$ possible
ones.

\subsection{Corner Chrominance Extraction}

\begin{figure}
  \centering
  \includegraphics{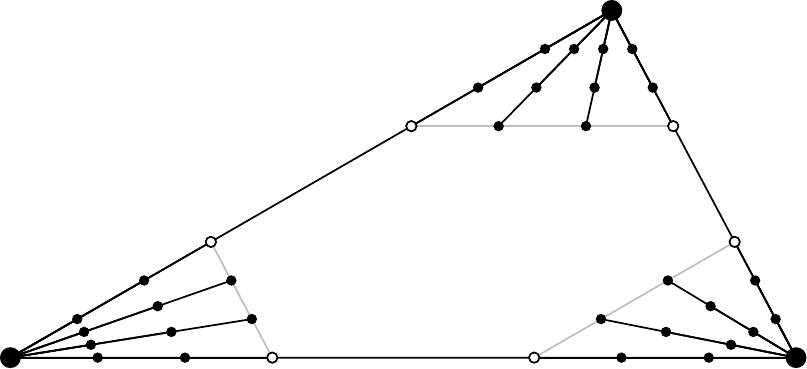}
  \caption{\label{fig:chrominance} Corner chrominance extraction. The
    gray segments define a limit for the size of the projected
    lines. The white points defining the extremities of those lines
    are positioned according to a fraction of the edge they belong
    to. In the above example the fraction is~$1/3$.}
\end{figure}

Figure \ref{fig:chrominance} illustrates the scheme for extracting a
feature vector from a keygraph. Basically, the extraction consists in
taking several internal segments and, for each one of them, to
calculate the mean chrominance of all pixels intersected by the
segment.

The chrominance values are obtained by converting the model to the HSV
color space and considering only the hue and saturation
components. The segments are obtained by evenly partitioning bundles
of lines projected from the vertices. Finally, the size of those
projected lines is limited by a segment whose extremities are points
in the keygraph edges.

This scheme is naturally invariant to rotation. Invariance to
brightness is ensured by the fact that we are considering only the
chrominance and ignoring the luminance. Finally, the invariance to
scale is ensured by the fact that the extremities mentioned above are
positioned in the edges according to a fraction of the size of the
edge that they belong to, and not by any absolute value.

\subsection{Results with Delaunay Triangulation}

We could not use, during the classification phase, the same keygraph
detector we used during the training phase: it does not reduce enough
the keygraph set size for real-time performance. We use an alternative
detector that gives us a smaller subset of the set the training
detector would give.

This alternative detector consists in selecting thick scalene
triangles from a \emph{Delaunay triangulation} of the keypoints. A
triangulation is a good source of triangles because it covers the
entire convex hull of the keypoints. And the Delaunay triangulation,
in particular, can be calculated very efficiently, for example with
the $\Theta(n \lg n)$ Fortune \cite{fortune:p2ascg-313-1986}
algorithm.

\begin{figure}
  \centering
  \includegraphics[scale=0.5]{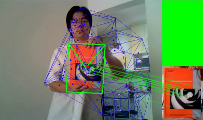}
  \includegraphics[scale=0.5]{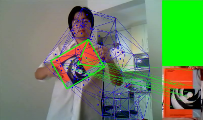}
  \includegraphics[scale=0.5]{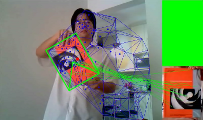}\\
  \vspace*{1pt}\includegraphics[scale=0.5]{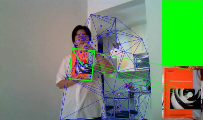}
  \includegraphics[scale=0.5]{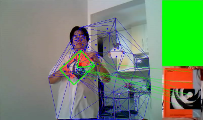}
  \includegraphics[scale=0.5]{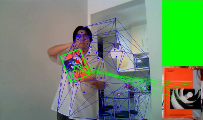}\\
  \vspace*{1pt}\includegraphics[scale=0.5]{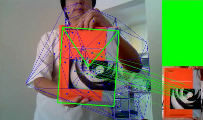}
  \includegraphics[scale=0.5]{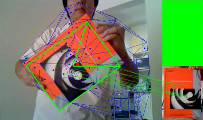}
  \includegraphics[scale=0.5]{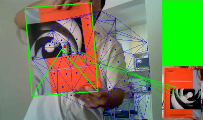}
  \caption{\label{fig:result} Results showing object detection robust
    to scale and rotation.}
\end{figure}

Figure \ref{fig:result} shows some resulting screenshots. A full video
can be seen at
\begin{center}
  \texttt{http://www.vision.ime.usp.br/\~{}mh/gbr2009/book.avi}.
\end{center}

\section{Conclusion}

We presented a new framework for keypoint-based object detection that
consists on classifying keygraphs. With an implementation of this
framework, where the keygraphs are thick scalene triangles, we have
shown successful results for real-time detection after training with a
single image.

The framework is very flexible and is not bounded to an specific
keypoint detector or keygraph detector. Therefore, room for
improvement lies on both the framework itself and the implementation
of each one of its steps. We are currently interested in using more
sophisticated keygraphs and in adding the usage of \emph{temporal
  information} to adapt the framework to \emph{object tracking}.

Finally, we expect to cope with 3D poses (i.e. out-of-plane rotations)
by incorporating aditional poses to the training set. These advances
will be reported in due time.

\section*{Acknowledgments}

We would like to thank FAPESP, CNPq, CAPES and FINEP for the support.

\nocite{}

\bibliographystyle{plain}
\bibliography{keygraph}

\end{document}